\documentclass[journal]{IEEEtran}  % Comment this line out if you need a4paper
\usepackage{amsmath}
\usepackage{amssymb}
\usepackage[colorinlistoftodos]{todonotes}
\usepackage{hyperref}
\usepackage{multirow}
\usepackage{booktabs}
\usepackage[colorinlistoftodos]{todonotes}
\usepackage{siunitx}
\usepackage{url}
\usepackage{import,bm}
\usepackage[super]{nth}
\usepackage{romannum}
\usepackage{accents}
\usepackage{verbatim}
\usepackage{tikz}
\usetikzlibrary{automata, positioning, arrows, shapes.geometric, arrows.meta, positioning, calc, intersections, spath3}
\usepackage{tkz-euclide}
\usepackage{subcaption}
\usepackage{graphicx}
\usepackage{cite}
\usepackage{makecell}
\usepackage{caption}
\usepackage{ragged2e}
\usepackage{mathtools}
\usepackage{tikz}
\usetikzlibrary{automata, positioning, arrows, shapes.geometric, arrows.meta, positioning, calc, intersections, spath3}
\usepackage{tkz-euclide}
\usepackage{pifont}
\usepackage{lipsum}

\newcommand{\claim}[1]{\noindent\textbf{#1}}

\newcommand{\etal}{\textit{et al}.}

\title{Real-World Reinforcement Learning with MPC Scaffolding for Dexterous Manipulation}

\author{Emek Barış Küçüktabak$^{1,*}$, Karankumar Patel$^{1,\dagger,*}$, Zhaodong Yang$^{1,2,*}$, Jinda Cui$^{1}$, Kazuhiro Sasabuchi$^{1}$, Jun~Takamatsu$^{1}$

\thanks {$^*$ equal contribution}
\thanks {$^1$ Honda Research Institute USA, San Jose, CA, USA}
\thanks {$^2$ Georgia Institute of Technology, Atlanta, GA, USA (Work done during an internship at HRI)}
\thanks {$^\dagger$ Current address: Skild AI, San Mateo, CA, USA (Work done while at HRI)}
}

\begin{document}
\maketitle
\thispagestyle{empty}
\pagestyle{empty}

\begin{abstract}
Real-world reinforcement learning (RL) offers a promising route to dexterous manipulation policies that can adapt directly from physical interaction, but learning is hindered by inefficient early exploration and costly failures. We propose a framework that uses sampling-based model predictive control (MPC) as scaffolding for real-world dexterous RL, providing structured prior experience and task-directed guidance during learning without human demonstrations or corrective actions. A small set of MPC trajectories is first used to populate an offline replay buffer and to pretrain the actor and critic. During online learning, MPC intermittently guides data collection while an off-policy Soft Actor-Critic learner trains from both prior MPC experience and newly collected physical interaction, with control gradually transitioning to the learned policy.  
On continuous in-hand rotation with a 16-DoF Allegro hand, initialized from 20 MPC trajectories collected in 12 minutes on hardware, the policy reaches 100\% success after 7 minutes of online RL, with about three object drops on average during training. After 20 minutes of online learning, the policy achieves more than five times the rotation speed of the MPC controller and completes 1000 consecutive rotations without a drop. Ablations show complementary benefits from MPC-based pretraining, retained MPC experience, and online MPC guidance, while additional experiments demonstrate rapid adaptation to new object geometries and successful goal-conditioned reorientation.\\
Project webpage: \href{https://real-world-rl-with-mpc.github.io}{real-world-rl-with-mpc.github.io}

% \begin{IEEEkeywords}
% keywords 10, keywords 20, keywords 30,
% \end{IEEEkeywords}

\end{abstract}
\section{Introduction}
\label{sec:Intro}

\begin{figure*}[t]
    \centering
    \includegraphics[width=0.9\textwidth]{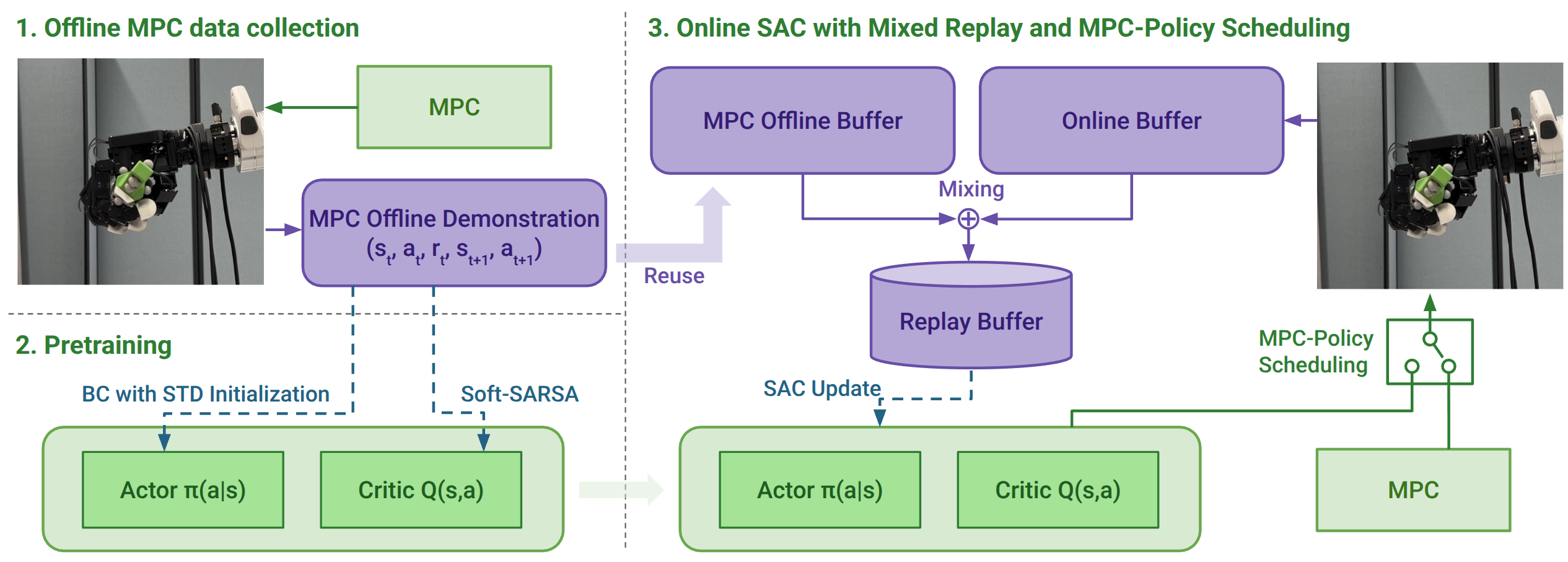}
    \caption{Overview of our framework. Stage 1 (Offline MPC Data Collection): Trajectories are collected via MPC. Stage~2 (Pretraining): The collected demonstrations pretrain the actor using behavior cloning with standard deviation initialization and the critic via Soft-SARSA. Stage 3 (Online SAC with Mixed Replay and MPC-Policy Scheduling): The actor-critic network is fine-tuned with SAC using a mixed replay buffer combining offline demonstrations and online transitions. An MPC-policy scheduling mechanism dynamically alternates control between the learned policy and MPC for guided exploration.}
    \label{fig:overview}
\end{figure*}

Multi-fingered robotic hands enable contact-rich behaviors such as in-hand reorientation, regrasping, and dexterous tool manipulation~\cite{openai2020dexterous,kedia2026simtoolreal}. However, their high-dimensional action spaces and complex contact dynamics make reliable control challenging. For practical deployment, dexterous policies must not only acquire such behaviors but execute them reliably over repeated operation. This motivates learning methods that can acquire robust dexterous skills efficiently on physical systems.

Existing approaches commonly obtain task-relevant dexterous behavior from human demonstrations~\cite{zhao2023learning,qin2022onehand,xu2025dexumi} or from policies trained in simulation~\cite{openai2020dexterous,handa2023dextreme,kedia2026simtoolreal, qi2023hora, chen2023visual}. Multi-fingered teleoperation can require specialized interfaces and substantial operator effort, while simulation-trained policies must contend with discrepancies in contact dynamics, friction, and hardware and object properties. Real-world reinforcement learning (RL) offers a complementary route: rather than requiring a simulator to capture all relevant physical effects, it allows the policy to improve directly from physical interaction and adapt to the dynamics actually encountered by the robot.

Recent advances in offline-to-online RL~\cite{lee2021offline, kostrikov2022offline, nair2021awacacceleratingonlinereinforcement} and RL fine-tuning\cite{Rajeswaran-RSS-18, mark2024policyagnosticrloffline}
have improved the online sample efficiency of robotic learning by leveraging previously collected data. Beyond using prior data for policy initialization, methods such as RLPD~\cite{ball2023rlpd} and SERL~\cite{luo2024serl} further improve sample efficiency by continuing to incorporate prior data throughout online learning. However, algorithmic efficiency alone does not ensure that the experience collected on hardware is informative, especially for high-dimensional dexterous manipulation. A policy that has just started training has a very small probability of producing coordinated actions that maintain contact while making meaningful task progress. Much of the early interaction can therefore consist of uninformative motion or rapid failures such as object drops. In the real world, these failures are particularly costly because they consume interaction time and can require physical resets.

One effective approach to improving real-world exploration is to supplement the learned policy with task-relevant guidance. Related ideas have appeared in Guided Policy Search~\cite{levine2013gps}, trajectory-optimization-based dexterous policy learning~\cite{kumar2016dexterous}, and Jump-Start RL~\cite{uchendu2023jsrl}, which use structured controllers or guide policies to bootstrap learning. HIL-SERL provides such guidance through human interventions for gripper-based manipulation, achieving impressive results on challenging tasks~\cite{luo2025hilserl}. Other recent work, including RL-100~\cite{lei2026rl100} and Ankile~\etal\cite{ankile2025residualoffpolicyrlfinetuning}, achieves strong real-world performance by starting from policies trained on human demonstrations and subsequently improving them through reinforcement learning. However, these methods rely on human-provided demonstrations or interventions to obtain task-relevant behavior. This dependence becomes particularly limiting for tasks where suitable teleoperation is impractical or unavailable, as can occur in multi-fingered manipulation.

Task-relevant exploration can alternatively be obtained from a model rather than a human. Khandate~\etal~use sampling-based planning to construct reset-state distributions for dexterous RL~\cite{khandate2023sampling}, and Li~\etal~convert MPC predictions into tracking rewards for humanoid locomotion and manipulation~\cite{li2026mpcrl}. In both, the model shapes rewards or initial states while the policy remains the only controller acting on the system; both therefore rely on large-scale simulation training and deploy to hardware without adaptation, leaving performance subject to sim-to-real discrepancies. In parallel,  PDDM learns dynamics from real-world interaction and uses sampling-based MPC for dexterous manipulation~\cite{nagabandi2020pddm}, while DROP~\cite{li2024drop} and Hess et al.~\cite{hess2025sampling} demonstrate sampling-based MPC directly on physical multi-fingered hands. In these approaches, MPC remains the controller at execution time, so task performance is ultimately determined by the learned or specified model and the online planning procedure rather than by a policy that improves through real-world interaction.

Sampling-based MPC and real-world RL offer complementary advantages. Although suboptimal, MPC can provide sufficiently good task-directed behavior for learning, and rollout constraints can steer it away from failure-prone trajectories, reducing object drops and human resets. RL can then build on this guided experience, optimize the task objective directly through real-world interaction, and learn a policy that outperforms the planner that guided it. The challenge is to make MPC a scaffold rather than a ceiling, structuring early exploration without capping the learned policy at the planner's performance.

% Building on these ideas, we ask whether a fixed sampling-based MPC controller can serve as a practical source of prior experience and temporary online guidance for modern off-policy RL directly on a dexterous hand, while ultimately being removed at deployment. We investigate this question by using MPC-generated trajectories both to initialize learning and to provide task-relevant exploration during early online interaction.

This motivates our central question: can sampling-based MPC provide structured prior experience and temporary task-directed guidance that enable efficient real-world dexterous RL while limiting costly human resets? Before online interaction, a small set of MPC trajectories initializes the replay buffer and is used to pretrain the actor and critic. During real-world training, the robot collects both MPC-guided and policy-generated experience, while control gradually transitions toward the learned policy. The off-policy learner trains from the initial MPC data together with newly collected physical experience. In this way, MPC serves as temporary scaffolding for learning, reducing reliance on uninformative and failure-prone early policy exploration, while RL % reinforcement learning 
adapts the policy to the dynamics encountered on the real system and learns a policy that executes independently of MPC.

We evaluate the proposed method primarily on continuous in-hand object rotation. After 12 minutes of MPC interaction to collect the 20-trajectory initialization dataset, the robot learns the rotation behavior in approximately $7$ minutes of online interaction and reaches its saturated rotation speed after approximately $15$ minutes, while incurring only $3$ object drops on average during training. In comparison, standard RL does not achieve a successful rotation within $45$ minutes and incurs $181$ object drops. Our trained policy performs $1000$ consecutive rotations without a single drop in close to two hours of uninterrupted execution. Through comparisons and ablations on the in-hand rotation task, we isolate the effects of MPC-based pretraining, online MPC guidance, and replay-buffer design. Starting from a policy trained on one object geometry, the policy transfers successfully to new geometries, and continued real-world learning substantially improves its execution speed. Additionally, we demonstrate that the proposed method extends beyond continuous rotation to goal-conditioned precision reorientation.

% The main contributions of this work are:
% \begin{itemize}
%     \item We introduce a real-world RL pipeline that uses sampling-based MPC both to initialize learning and to provide task-relevant online exploration for contact-rich multi-fingered manipulation, reducing reliance on random exploration or continuous human intervention.
%     \item We incorporate MPC experience through three complementary roles: retained prior replay, actor–critic pretraining, and temporary online guidance—and experimentally isolate the contribution of each role to learning efficiency and physical reset burden.
%     \item We demonstrate rapid and low-intervention real-world learning on dexterous in-hand manipulation, including learning rotation within minutes, adaptation to changes in object size, and $1000$ consecutive rotations over more than $110$ minutes without a drop, together with ablations of the individual components of the method.
% \end{itemize}

In summary, the main contributions of this work are:
\begin{itemize}
    \item We propose a real-world RL framework that uses sampling-based MPC for prior replay, actor-critic pretraining, and temporary online guidance in contact-rich multi-fingered dexterous manipulation.
    % after which control transitions entirely to the learned policy without requiring human demonstrations or online human intervention.
    
    \item Through controlled ablations,  we show that the three uses of MPC are complementary: replay reduces object drops, pretraining accelerates early learning and further lowers the reset burden, and online guidance improves learning speed and consistency.
    
    \item We demonstrate real-world learning within minutes, followed by $1000$ consecutive rotations without a drop, as well as adaptation to new object geometries and goal-conditioned reorientation.
\end{itemize}

\section{Method}
\label{sec:method}

\subsection{Overview}
\label{sec:method_overview}

Our framework uses sampling-based MPC as a scaffold for an off-policy learner, with the learned policy progressively replacing MPC during real-world interaction. As illustrated in Fig.~\ref{fig:overview}, training consists of three stages. First, a small set of task-directed trajectories is collected using a sampling-based MPC controller. Second, these trajectories are used to pretrain the actor and critic of a Soft Actor-Critic (SAC) agent~\cite{haarnoja2018soft}. Third, we perform online RL using SAC through real-world interaction, while retaining the initial MPC data in an offline replay buffer and intermittently using MPC to guide online data collection. The probability of policy execution is gradually increased until the learned policy fully replaces MPC. Both MPC and the learned policy produce commands through the same joint-level hand-control interface, allowing the resulting transitions to be used by the same off-policy learner regardless of which controller generated them.

% MPC therefore contributes to learning through both the initial dataset and early online interaction, while the policy continues to improve from real-world experience using the off-policy RL objective and ultimately executes independently of the planner at deployment. Both MPC and the learned policy produce commands through the same joint-level hand-control interface, allowing the resulting transitions to be used by the same off-policy learner regardless of which controller generated them.

% The MPC controller is used as a fixed guidance module. It maintains a MuJoCo digital twin synchronized with the measured robot and object state and optimizes short-horizon control trajectories using the cross-entropy method (CEM)~\cite{li2024drop}. For continuous rotation, the MPC uses a low-dimensional motion primitives, together with joint-level residual sampling that allow the sampled motions to adapt to the current hand-object configuration. Candidate trajectories are evaluated according to a task-specific MPC objective, and the leading action of the optimized trajectory is executed in receding-horizon fashion. We use the primitive-informed sampling-based MPC formulation of~\cite{kucuktabak2026primitive} and refer to that work for the complete planner formulation.

The MPC controller is used as a fixed guidance module. It maintains a MuJoCo digital twin synchronized with the measured robot and object state and optimizes short-horizon control trajectories using the cross-entropy method (CEM)~\cite{kobilarov2012cross, li2024drop, kucuktabak2026primitive}. %For continuous rotation, 
The controller uses low-dimensional motion primitives, together with joint-level residual sampling that allows the sampled motions to adapt to the current hand-object configuration. Candidate trajectories are evaluated according to a task-specific MPC objective, and the leading action of the optimized trajectory is executed in receding-horizon fashion. The MPC formulation itself is adopted from~\cite{kucuktabak2026primitive}; we refer to that work for the primitive construction and complete planner formulation, and use the controller here as a fixed source of prior experience and online guidance.

We formulate each manipulation task as a continuous-control MDP, where the observation $o_t$ contains robot proprioception together with task-relevant object information, and the action $a_t$ specifies joint-level commands for the dexterous hand. Rewards are defined from task progress and terminal success or failure. The task-specific observation, action, and reward definitions are provided in Sec.~\ref{sec:experiments}.

\subsection{Offline MPC Data and Pretraining}
\label{sec:mpc_initialization}

Before online learning, we collect a small dataset of trajectories using the MPC controller,
$\mathcal{D}_{\mathrm{MPC}}$. This dataset serves two purposes. First, all collected trajectories are retained as a fixed offline replay buffer for online learning (Sec.~\ref{sec:mixed_replay}). Second, the dataset is used to pretrain the SAC actor and critic before online learning begins. Each trajectory contains transitions
\begin{equation}
    (s_t,a_t,r_t,s_{t+1},a_{t+1}),
\end{equation}
where $s_t$ denotes the policy observation, $a_t$ the hand command executed on the system, and $r_t$ the task reward. To align with online learning, the transitions are labeled using the same RL reward used during online learning rather than the MPC %planning 
objective.

Our pretraining framework consists of two sequential stages: behavior cloning (BC) to initialize the actor, followed by a Soft-SARSA phase to pretrain the critic.

\subsubsection{Actor Pretraining}
\label{sec:actor_initialization}

Unlike standard behavior cloning, our approach requires consideration of the standard deviation (std) head within the stochastic actor network. Instead of co-training the mean head and the std head, we implement a constant base std initialization. This is achieved by directly setting the weights of the last layer of the std head to 0 and its bias to the std target. %This initialization strategy achieves better mean policy performance than co-training both heads simultaneously. 
To optimize the BC objective on sub-optimal MPC demonstrations, we use two additional choices to improve actor initialization. First, to prioritize high-quality demonstrations, we implement a weighted episode sampling strategy, wherein the sampling probability $P(i)$ of each trajectory $\tau_i$ is modulated by a softmax distribution over its episodic return. % For tasks with high variance in demonstration quality, we further restrict this distribution by assigning zero probability to all but the top-$K$ performing episodes.
Second, to mitigate action saturation during training, we apply epsilon clipping to the target actions.  %Specifically, the target action $a_{target}$ is derived by applying an inverse hyperbolic tangent ($\text{arctanh}$) transformation to the raw demonstration action $a^{demo}$, which is explicitly bounded by a predefined $\epsilon$ margin:
  %$$a_{target} = \text{arctanh}(\text{clip}(a^{demo}, -1+\epsilon, 1-\epsilon))$$
By integrating these bounded targets with our weighted sampling strategy, we formulate the behavior cloning objective as a supervised regression task. We minimize the Mean Squared Error (MSE) between the actor network's pre-tanh mean prediction, denoted as $\mu_\theta(s)$, and the transformed target action:

\begin{equation}
\begin{aligned}
  \mathcal{L}_{\text{BC}}(\theta) &= \mathbb{E}_{\tau_i \sim P, (s, a^{\text{demo}}) \sim \tau_i} \Big[ \Big\Vert \mu_\theta(s) - \\
  &\quad \text{arctanh}(\text{clip}(a^{\text{demo}}, -1+\epsilon, 1-\epsilon)) \Big\Vert^2 \Big].
\end{aligned}
\end{equation}

By training the mean head and actor trunk via MSE regression while directly initializing a constant base std, 
the policy accurately captures demonstration behaviors, while preserving necessary and regularized stochasticity.
This also ensures the pretrained policy provides log-probabilities, $\log \pi_{BC}$, which act as the entropy term in the Soft-SARSA Bellman target of the subsequent critic pretraining phase.

\subsubsection{Critic Pretraining}
\label{sec:critic_initialization}
To pretrain the critic network, we use a Soft-SARSA objective that augments the standard SARSA Bellman target with entropy regularization. The target value is computed as:
\begin{equation}
\label{eq:soft_sarsa}
\begin{aligned}
y_{t} &= r_{t} + \gamma \big[ \min_{j}\overline{Q}_{j}(s_{t+1},a_{t+1}^{\text{demo}}) -  \alpha_{\text{eff}} \log \pi_{\text{BC}}(a_{t+1}^{\text{demo}}, s_{t+1}) \big].
\end{aligned}
\end{equation}
In this formulation, $r_t$ represents the scalar reward observed at timestep $t$, sampled directly from $\mathcal{D}_{\mathrm{MPC}}$, $\gamma \in [0, 1)$ is the discount factor, and $a_{t+1}^{\text{demo}}$ is the subsequent action executed in the demonstration data. 
To stabilize the critic target, at each update we randomly sample two critic indices from an ensemble of 10 target critic networks and take the minimum of their predictions, written as $\min_{j}\overline{Q}_{j}(s_{t+1},a_{t+1}^{\text{demo}})$ in Eq.~\ref{eq:soft_sarsa}. The same critic-ensemble scheme is used throughout online fine-tuning.
% To stabilize Q-learning updates and suppress baseline function approximation errors, the term $\min_{j}\overline{Q}_{j}(s_{t+1},a_{t+1}^{\text{demo}})$ computes the minimum over an ensemble of 10 target critic networks. This architectural choice is maintained throughout the online fine-tuning phase.
Finally, the term $\alpha_{\text{eff}} \log \pi_{\text{BC}}(a_{t+1}^{\text{demo}}, s_{t+1})$ acts as a maximum entropy bonus. The coefficient $\alpha_{\text{eff}}$ serves as a fixed, effective temperature parameter, scaling the log-probability of the demonstration action under the frozen BC actor. This ensures architectural and mathematical consistency with the entropy-augmented objectives of the subsequent online SAC phase. 

% Furthermore, this Soft-SARSA objective is highly extensible and can readily incorporate a conservative Q-learning (CQL)~\cite{CQL} regularization term. By penalizing out-of-distribution actions while lower-bounding Q-values against the true returns of the demonstration data, this term provides an effective safeguard against the offline-to-online performance dip. This Soft-SARSA formulation is deliberately designed to align with the subsequent online SAC updates. 

% Empirically, pretraining the critic with this approach yields superior downstream performance compared to naively training the SAC agent directly on the offline dataset.

\subsection{Online SAC with Mixed Replay}
\label{sec:mixed_replay}

Following pretraining, the actor and critic networks are continuously optimized online via SAC. Inspired by RLPD~\cite{ball2023rlpd} and HIL-SERL~\cite{luo2025hilserl}, we employ a mixed replay strategy. We maintain a fixed offline replay buffer containing the initial MPC trajectories and an online replay buffer containing all transitions collected during real-world interaction. During this online phase, transitions generated by both MPC-guided and policy-controlled execution are stored in the same online buffer and are treated identically by the off-policy learner.

% Each minibatch contains $30\%$ samples from the fixed MPC dataset and $70\%$ from newly collected online experience. In comparison, HIL-SERL~\cite{luo2025hilserl} samples $50\%$ of each training batch from its demonstration/intervention buffer. We use a lower prior-data ratio because the MPC trajectories provide useful task-directed behavior but are generally suboptimal relative to the performance desired by the learned policy. The smaller offline fraction preserves this useful experience during early learning while allowing newly collected online data to increasingly drive policy improvement beyond the MPC behavior.

HIL-SERL~\cite{luo2025hilserl} samples $50\%$ of each training batch from its demonstration/intervention buffer. In our setting, each minibatch draws $30\%$ of its samples from the fixed MPC dataset and $70\%$ from newly collected online experience. We use a smaller prior-data fraction because the MPC trajectories provide useful task-directed behavior but are generally suboptimal relative to the performance expected of the learned policy. This split retains the MPC experience during early learning while allowing online experience to drive policy improvement beyond the MPC behavior. Additional implementation details, and training hyperparameters for pretraining and online SAC are listed on the project webpage.

% Both MPC-generated and policy-generated transitions are optimized using the standard SAC objective; MPC actions are not used as supervision targets during online learning. Thus, MPC shapes the experience available to the learner without constraining the policy to imitate the planner.

\subsection{MPC-Policy Scheduling}
\label{sec:controller_schedule}

During online learning, MPC guidance is gradually reduced as training progresses. The probability of executing the learned policy is
\begin{equation}
\label{eq:policy_schedule}
p_{\pi}(k)
=
p_0
+
(1-p_0)
\min\left(\frac{k}{K},1\right),
\end{equation}
where $k$ is the number of online training environment steps, $p_0$ is the initial probability of policy execution, and $K$ specifies the number of steps over which this probability increases linearly to one. We use $p_0=0.5$ and $K=6000$ environment steps (roughly 10 minutes of wall-clock time). After this point, MPC guidance is no longer used during interaction.

Rather than selecting between MPC and the policy at every environment step, the controller is sampled once every $L=40$ steps and used for the entire chunk. Chunk-level switching preserves temporally coherent finger-motion sequences and avoids rapid alternation between independently generated MPC and policy commands. %during contact-rich manipulation.
MPC and policy actions are not blended; only the selected controller is executed.

\section{Experiments and Results}
\label{sec:experiments}

We evaluate the proposed framework through real-world dexterous manipulation experiments. We first use continuous in-hand rotation to study each component of our proposed pipeline and compare the learned policy against direct MPC execution and an RL policy trained entirely in simulation and deployed directly on hardware. We then evaluate adaptation of the learned policy to objects with different geometries. Finally, we apply the same learning framework to goal-conditioned object reorientation to demonstrate applicability beyond continuous rotation. Videos of the experiments are provided in the supplementary video and on the project webpage.

\subsection{Experimental Setup}
\label{sec:experimental_setup}

\begin{figure}[t]
\centering
\includegraphics[width=\columnwidth]{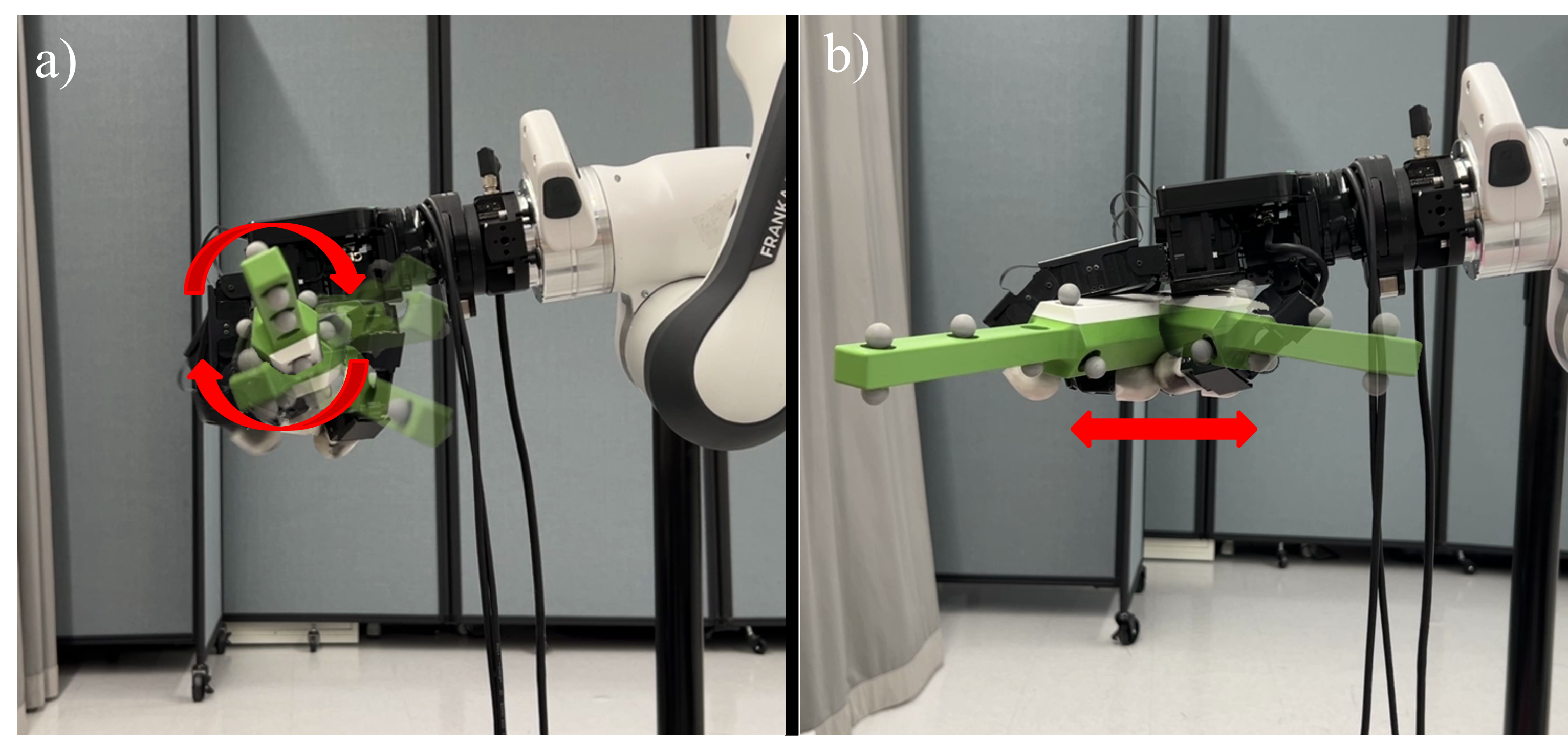}
\caption{Experimental setup. (a) continuous in-hand rotation of a hexagonal object. (b) goal-conditioned reorientation toward a commanded object heading.}
\label{fig:exp_setup}
\end{figure}

\paragraph{Hardware platform}
Experiments are conducted using a 16-DoF Allegro hand (Wonik Robotics) mounted on a Franka FR3 arm (Franka Robotics). %, which remains fixed throughout the experiments.
The learned policy commands joint position targets for the Allegro hand at 10 Hz through the learning framework described in Sec.~\ref{sec:method}. Object pose is measured using external motion capture (OptiTrack) and is provided to both the policy observation and the synchronized MuJoCo model used by the sampling-based MPC.

\paragraph{Continuous in-hand rotation}
Our primary task is continuous in-hand rotation using a hexagonal object with an across-flats (AF) dimension of 45~mm. The objective is to repeatedly rotate the object about its primary axis while maintaining it within the hand. We represent rotation progress using the rotation angle $\psi_t$, such that rotation in the desired direction increases $\psi_t$ while rotation in the opposite direction decreases it.

The policy observation is
\begin{equation}
o_t =
\left[
q_t^h,\,
p_t^{ee,o},\,
q_t^o,\,
\psi_t,\,
\psi_{t-1},\,
\dot q_t^h,\,
\dot p_t^{ee,o},\,
a_{t-1}
\right],
\end{equation}
where $q_t^h$ denotes the 16 hand joint positions, $p_t^{ee,o}$ is the object position relative to the end effector, $q_t^o$ is the object orientation, $\dot q_t^h$ and $\dot p_t^{ee,o}$ are finite-difference velocity estimates, and $a_{t-1}$ is the previously executed hand command. The action $a_t\in[-1,1]^{16}$ is mapped to joint position targets for the Allegro hand.

To avoid task-specific reward engineering, we use a simple RL reward defined only by physical task progress and terminal outcomes. For continuous rotation, we use
\begin{equation}
\label{eq:rotation_reward}
r_t =
\lambda_{\psi}
(\psi_t-\psi_{t-1})
+
\lambda_{\tau}
+
b_{\mathrm{succ}} I_{\mathrm{succ}},
\end{equation}
where $\lambda_{\psi}=20$ scales the rotation-progress reward, $\lambda_{\tau}=-0.3$ penalizes elapsed time, and $I_{\mathrm{succ}}$ is an indicator that equals one when a full rotation is completed and zero otherwise. The coefficient $b_{\mathrm{succ}}=200$ specifies the corresponding success bonus. When an object drop is detected, the reward is replaced by the terminal penalty $b_{\mathrm{drop}}=-50$.

% For the continuous-rotation experiments, we collect $20$ successful trajectories using the sampling-based MPC controller, each completing a $2\pi$ object rotation in the desired direction without a drop, before online learning. All trajectories are retained in the fixed offline replay buffer and used for critic pretraining, while the highest-return 10 trajectories are used for actor pretraining.

For the continuous-rotation experiments, before online RL we collect 20 trajectories using the sampling-based MPC controller, each completing a \(2\pi\) object rotation in the desired direction without a drop. The controller succeeds on all 20 collection attempts in approximately 12 minutes of physical interaction. All trajectories are retained in the fixed MPC replay buffer and used for actor-critic pretraining. Throughout the paper, reported online-learning times are measured from the start of SAC interaction after this MPC-data collection and pretraining stage.

To track learning progress during online training, we periodically evaluate the current policy over five policy-only episodes without MPC guidance. A trial is considered successful if the policy completes a full $2\pi$ rotation without dropping the object within the evaluation horizon of $90$ seconds. We report success rate to characterize the emergence of reliable rotation, rotation speed to measure improvement in execution performance, and the cumulative number of object drops during online training to quantify the physical reset burden of learning. These five-episode evaluations are used to track learning progress; extended policy evaluation is conducted separately over larger sets of trials, as described in Sec.~\ref{sec:rotation_final}.

\paragraph{Ablations}
We evaluate the contributions of the three uses of MPC in our framework:
retaining the initial MPC trajectories as offline replay data (\textbf{R}),
using these trajectories for actor-critic pretraining (\textbf{P}), and
using MPC to guide online interaction (\textbf{G}).
Table~\ref{tab:ablations} summarizes the six evaluated configurations.
All configurations use the same online SAC learner; the labels indicate
which of the three components are enabled.
These configurations enable controlled comparisons of pretraining
(Full method vs.\ No Pretrain), retained MPC replay
(No Pretrain vs.\ Guidance Only), and online MPC guidance
(Full method vs.\ No Guidance).

\begin{table}[t]
    \centering
    \caption{Ablation configurations. R: MPC replay; P: actor-critic
    pretraining; G: online MPC guidance.}
    \label{tab:ablations}
    \setlength{\tabcolsep}{6pt}
    \begin{tabular}{lccc}
        \hline
        Configuration & R & P & G \\
        \hline
        Full method   & \checkmark & \checkmark & \checkmark \\
        No Pretrain   & \checkmark & $\times$   & \checkmark \\
        Guidance Only & $\times$   & $\times$   & \checkmark \\
        No Guidance   & \checkmark & \checkmark & $\times$   \\
        Replay Only   & \checkmark & $\times$   & $\times$   \\
        RL Only       & $\times$   & $\times$   & $\times$   \\
        \hline
    \end{tabular}
\end{table}

\subsection{In-Hand Rotation Results}
\label{sec:rotation_results}

\subsubsection{Ablation Results}

\begin{figure*}[t!]
\centering
\includegraphics[width = 2.0\columnwidth]
{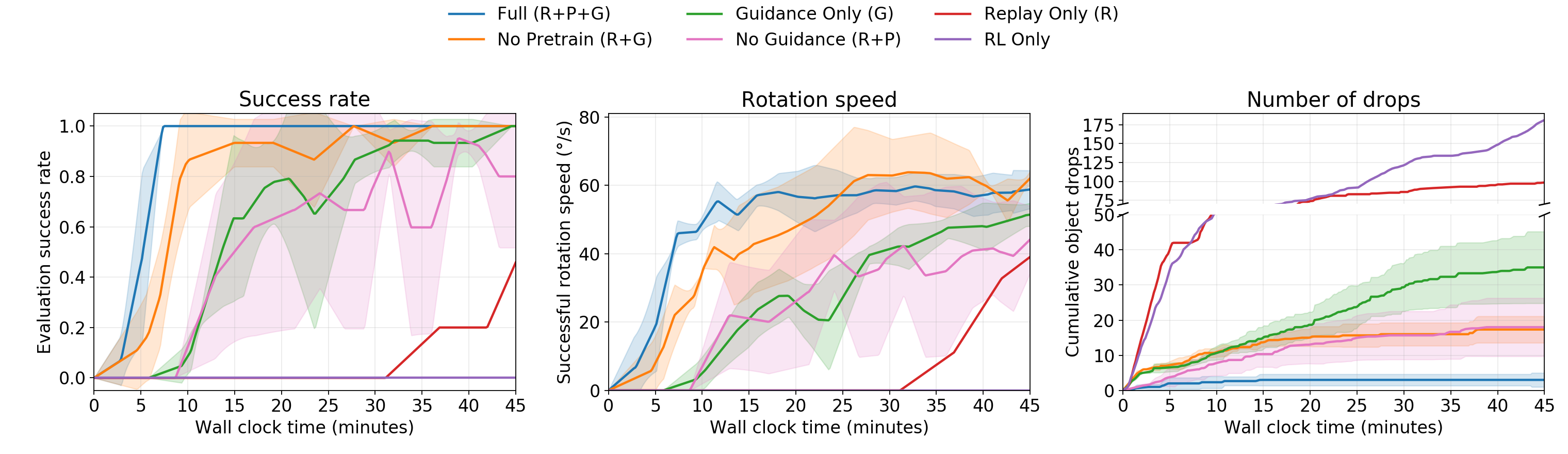}
\caption{Ablation results for continuous in-hand rotation. For the Full method, No Pretrain, Guidance Only, and No Guidance conditions, solid lines show the mean across three independent training runs and shaded regions indicate $\pm$ one standard deviation. Replay Only and RL Only are shown as single runs; because these conditions exhibited high object-drop rates, consecutive repeat testing was avoided to limit hardware risk.}
\label{fig:rotation_ablation}
\end{figure*}

% Figure~\ref{fig:rotation_ablation} compares the six training configurations in terms of evaluation success rate, rotation speed, and cumulative object drops. The \emph{Full method} (R+P+G) reaches its first 5/5 successful policy-only evaluation after approximately $7$ minutes of online interaction and approaches its saturated rotation speed after approximately $15$ minutes. Across three independent training runs, it incurs only approximately three object drops on average, indicating a low physical-reset and human-intervention burden during training.

Fig.~\ref{fig:rotation_ablation} compares the six configurations in Table~\ref{tab:ablations} in terms of evaluation success rate, rotation speed, and cumulative object drops.
 
\claim{The full method learns reliable rotation in minutes.} The \emph{Full method} (R+P+G) reaches its first 5/5 successful policy-only evaluation after $7.4 \pm 0$ minutes of online interaction and approaches its saturated rotation speed after approximately $15$ minutes, while incurring only approximately three object drops on average (1, 3 and 6) across three independent runs.

\claim{Pretraining speeds early learning and cuts drops.} In a separate policy-only evaluation, the pretrained policy reaches $140^\circ$ of rotation on average over 10 trials, but does not complete a full $2\pi$ rotation in any trial. This suggests that pretraining provides a useful task-relevant initialization, but is insufficient on its own to solve the rotation task. Comparing \emph{No Pretrain} (R+G) with the \emph{Full method} isolates actor-critic pretraining.  Pretraining reduces the time to the first 5/5 evaluation from approximately $16$ to $7$ minutes and the average number of drops from approximately $17$ to $3$. Although \emph{No Pretrain} eventually reaches $100\%$ success and a comparable rotation speed ($\approx 60^\circ$/s), the \emph{Full method} attains this level substantially earlier and with fewer drops.

\claim{Guidance alone enables learning.} \emph{RL Only} does not complete a successful rotation within the $45$-minute window and accumulates $181$ drops. In contrast, \emph{Guidance Only} (G) reaches its first 5/5 evaluation after $18.0 \pm 2.4$ minutes with $35.0 \pm 10.2$ drops over the same window.

\claim{MPC replay reduces drops but is insufficient on its own.} \emph{No Pretrain} (R+G) differs from \emph{Guidance Only} (G) only in retaining the offline MPC trajectories in replay. It reaches the first 5/5 evaluation after $15.5 \pm 8.6$ minutes, a comparable time, but accumulates only $17.3 \pm 3.8$ drops, roughly half as many. A similar pattern appears without guidance: \emph{Replay Only} (R) accumulates $98$ drops, compared with $181$ for \emph{RL Only}, although neither achieves reliable rotation within the $45$-minute window. These comparisons suggest that replay primarily reduces the physical reset burden rather than accelerating learning.

\claim{Guidance also makes early learning more consistent.} \emph{No Guidance} (R+P) isolates the effect of online guidance when replay and pretraining are already in place. It receives the same MPC replay and actor-critic pretraining as the \emph{Full method}, yet reaches its first 5/5 evaluation only after $25.8 \pm 9.4$ minutes across three runs and accumulates $18.0 \pm 8.3$ drops. In comparison, the \emph{Full method} reaches the 5/5 evaluation after $7.4$ minutes in all three runs and incurs approximately three drops on average. These results show that online guidance makes early learning substantially more consistent across runs, even when replay and pretraining are already provided.

\claim{The three roles are complementary.} MPC replay (R) preserves task-relevant prior experience, pretraining (P) provides a better starting point, and online guidance (G) improves the quality and consistency of newly collected interaction. Taken together, the ablations show that MPC's contribution is not reducible to a single initialization step. Indeed, their combination enables reliable and high-speed rotation substantially earlier while requiring far fewer object drops and physical resets.

\subsubsection{Extended Policy Evaluation}
\label{sec:rotation_final}

\begin{figure}[t!]
\centering
\includegraphics[width = 1.0\columnwidth]
{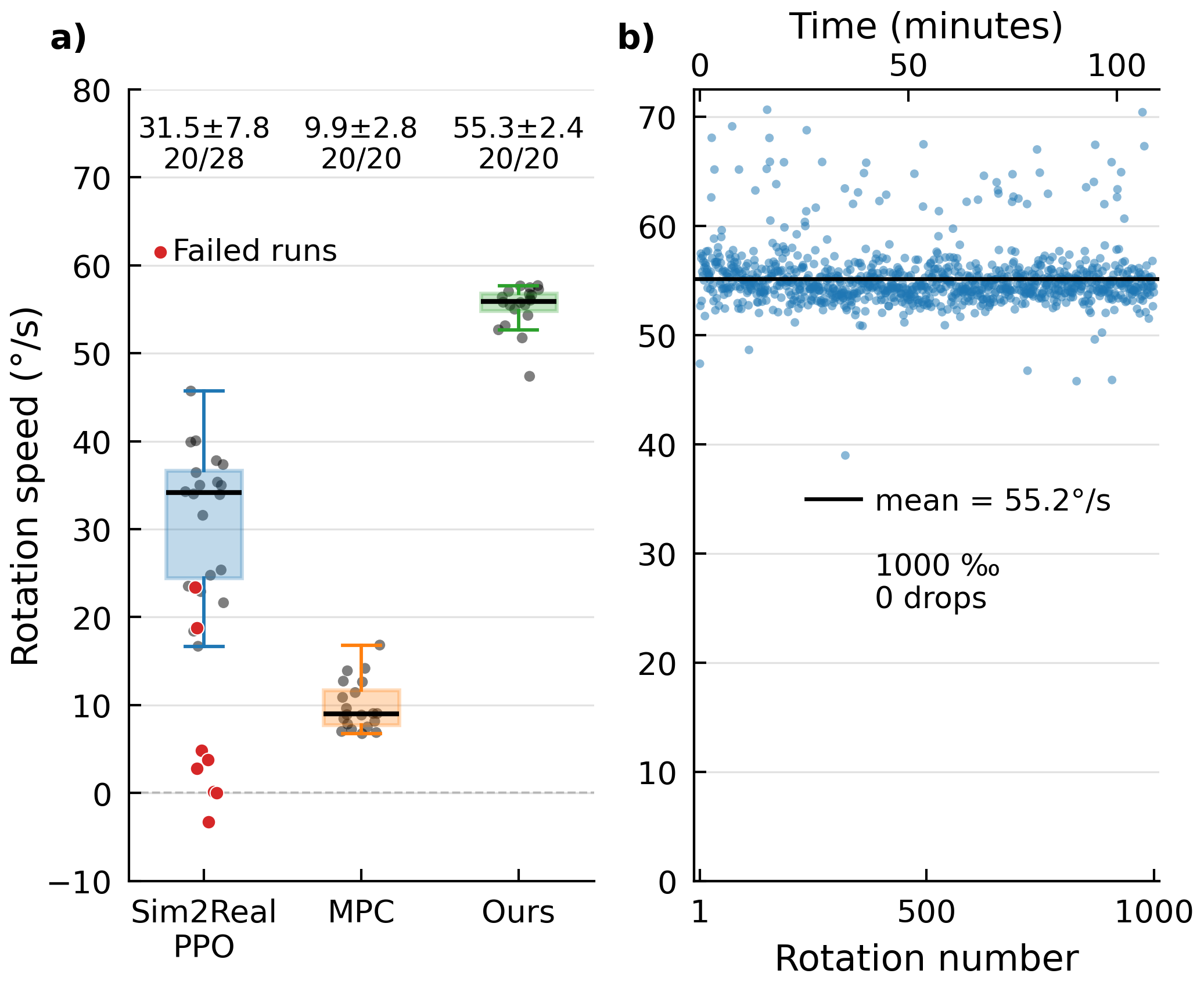}
\caption{Extended evaluation of the learned continuous-rotation policy. (a) Rotation speed for the sim-to-real PPO, direct MPC execution, and our learned policy. (b) Rotation speed over 1000 consecutive rotations using our policy.}
\label{fig:rotation_final}
\end{figure}

% We evaluate the policy obtained after 20 minutes of online learning against direct execution of the sampling-based MPC controller and a sim-to-real PPO~\cite{schulman2017proximalpolicyoptimizationalgorithms} baseline. The baseline is trained from scratch in MuJoCo with Stable-Baselines3~\cite{stable-baselines3}, using the same observation, action space, and control interface as our method and domain randomization over object mass, scale, friction, initial pose, and hand PD gains.

We evaluate the policy obtained after 20 minutes of online learning against direct execution of the sampling-based MPC controller and a PPO~\cite{schulman2017proximalpolicyoptimizationalgorithms} policy trained in simulation and deployed without adaptation. As a matched sim-to-real reference, the PPO policy is trained from scratch in MuJoCo with Stable-Baselines3~\cite{stable-baselines3}, using the same observation, action space, base MuJoCo model, and control interface as our method. Training uses 192 parallel environments for 15 million transitions with domain randomization over object mass, geometry scale, friction, initial object pose, and hand PD gains, together with observation and action noise. After training, the policy achieves 100/100 successful rotations in simulation.

\begin{figure}[t!]
\centering
\includegraphics[width = 1.0\columnwidth]
{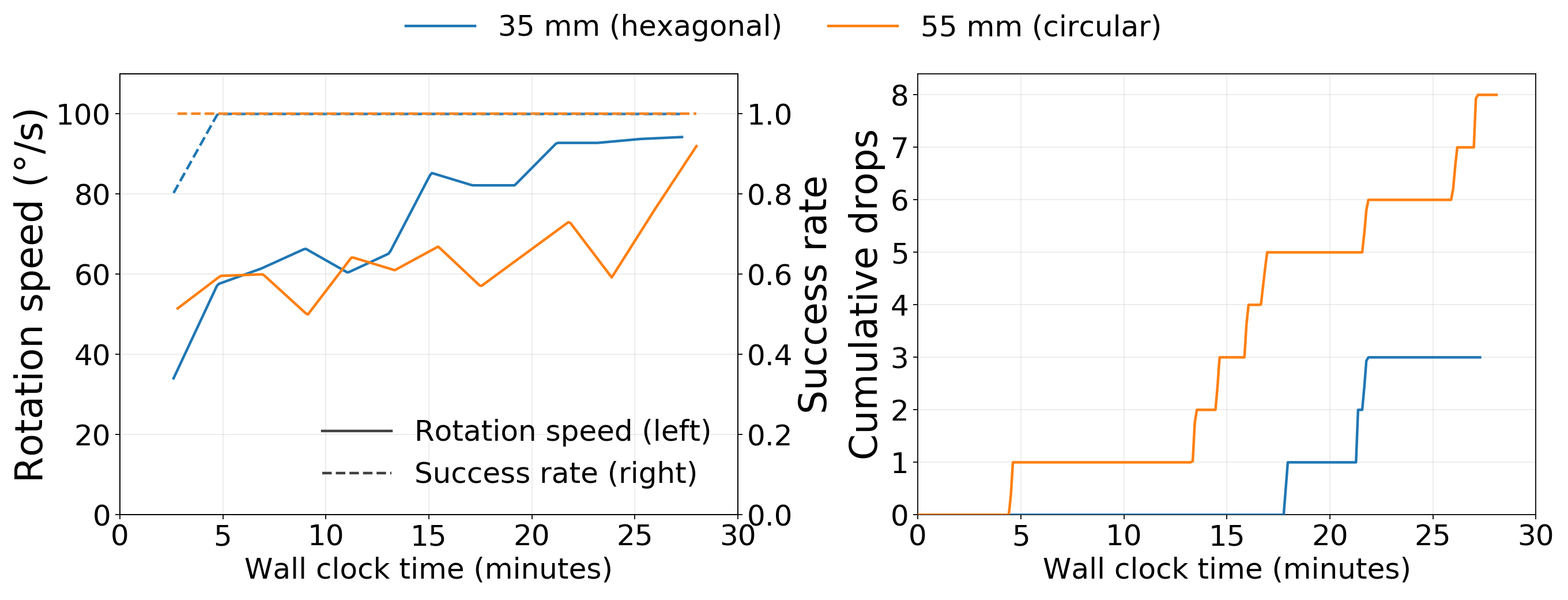}
\caption{Real-world adaptation to new object geometries. Policy-only evaluation success rate, successful-rotation speed, and cumulative object drops are presented.}
\label{fig:adaptation_training}
\end{figure}

% As shown in Fig.~\ref{fig:rotation_final}(a), we compare rotation speed over 20 successful rotations for each approach. The sim-to-real PPO baseline completes 20 of 28 evaluation trials ($71\%$) and achieves a mean successful-rotation speed of $31.5 \pm 7.8^\circ$/s. Direct MPC execution completes all 20 evaluated rotations and achieves $9.9 \pm 2.8^\circ$/s. Our policy likewise completes all 20 rotations and achieves $55.3 \pm 2.4^\circ$/s.

\claim{Outperforms the planner that guided it.} Fig.~\ref{fig:rotation_final}(a) compares rotation speed over 20 successful rotations for each controller. The sim-to-real PPO policy completes 20 of 28 trials (71\%) at $31.5 \pm 7.8^\circ$/s. Direct MPC execution completes all 20 rotations at $9.9 \pm 2.8^\circ$/s. Our policy completes all 20 at $55.3 \pm 2.4^\circ$/s, more than five times the speed of the planner that guided its training. The PPO policy succeeds in every simulated trial (100/100) and fails 8 of 28 on hardware, a transfer gap the hardware-trained policy does not incur.

% To further evaluate long-duration reliability, we execute our policy continuously for 1000 consecutive full rotations, as shown in Fig.~\ref{fig:rotation_final}(b). The policy completes all 1000 rotations without a single object drop, corresponding to more than $110$ minutes of uninterrupted physical execution, with a mean rotation speed of $55.2^\circ$/s. This extended evaluation demonstrates that the learned policy can sustain reliable, high-speed manipulation over prolonged repeated operation, supporting its suitability for applications that require consistent execution over many consecutive cycles.

\claim{1000 consecutive rotations without a drop.} To evaluate long-duration reliability, we execute the policy continuously for $1000$ consecutive full rotations (Fig.~\ref{fig:rotation_final}(b)). It completes all $1000$ without a single object drop, corresponding to approximately $110$ minutes of uninterrupted physical execution at a mean rotation speed of $55.2^\circ$/s (see supplementary video). This extended evaluation demonstrates that the learned policy can sustain reliable, high-speed manipulation over prolonged repeated operation, supporting its suitability for applications that require consistent execution over many consecutive cycles.

These results also highlight the complementary roles of planning and learning in our framework. Sampling-based MPC provides task-directed and reliable manipulation behavior, completing all evaluated rotations without a drop, although its direct execution is substantially slower than the learned policy. Using this reliable controller as a scaffold to initialize learning and guide early real-world interaction helps keep exploration away from failure-prone behaviors, contributing to the small number of object drops observed during training. SAC then improves from the resulting physical experience and ultimately executes independently of the planner. The resulting policy combines substantially faster execution than MPC with reliable repeated operation.

\subsection{Adaptation to Different Object Geometries}
\label{sec:adaptation}

% In practical settings, a policy may need to adapt to object geometries that differ from the one it was trained on. We therefore investigate whether a policy learned on one object geometry provides a useful initialization for manipulation of new geometries, and whether it can subsequently be specialized through additional real-world interaction without recollecting offline MPC data or repeating pretraining. Starting from the policy trained on the nominal 45~mm across-flats hexagonal object, we adapt the policy separately to a smaller 35~mm across-flats hexagonal object and a larger 55~mm-diameter circular object. During adaptation, no new offline MPC dataset is collected, and the original offline MPC replay buffer is not used. Learning therefore proceeds from the previously learned policy using only newly collected online experience, while online MPC guidance is retained to provide task-directed exploration. The observation, action space, and rotation reward remain unchanged.

In practical settings, a policy may need to adapt to object geometries that differ from the one it was trained on. We investigate whether a policy learned on one geometry provides a useful initialization for new geometries, and whether continued real-world learning with online MPC guidance can specialize it without recollecting offline MPC data or repeating pretraining. Starting from the policy trained on the nominal 45~mm across-flats hexagonal object, we adapt the policy separately to a 35~mm across-flats hexagonal object and 55~mm-diameter circular object. For each new object, we update only the corresponding geometry in the MPC model, avoiding the cost of collecting a new offline MPC dataset and repeating pretraining. Online MPC guidance provides task-directed exploration on the new geometry without any additional offline data collection or pretraining. The original replay buffer, whose transitions reflect the nominal geometry, is not reused, so learning proceeds from the existing policy using only newly collected online experience. The observation, action space, and rotation reward remain unchanged.

Fig.~\ref{fig:adaptation_training} shows that continued real-world learning progressively improves manipulation performance on both new objects. For the 35~mm hexagonal object, the first policy-only evaluation achieves 80\% success and subsequent evaluations reach 100\%, while for the 55~mm circular object, the transferred policy achieves 100\% success from the first evaluation. More importantly, rotation speed shows a sustained upward trend over the course of adaptation for both objects, indicating that the policy continues to specialize to each new geometry and improve its execution performance with additional real-world interaction. Adaptation also incurs relatively few resets, with only three object drops for the 35~mm object and eight for the 55~mm object over the full adaptation runs.

\begin{figure}[t!]
\centering
\includegraphics[width = 1.0\columnwidth]
{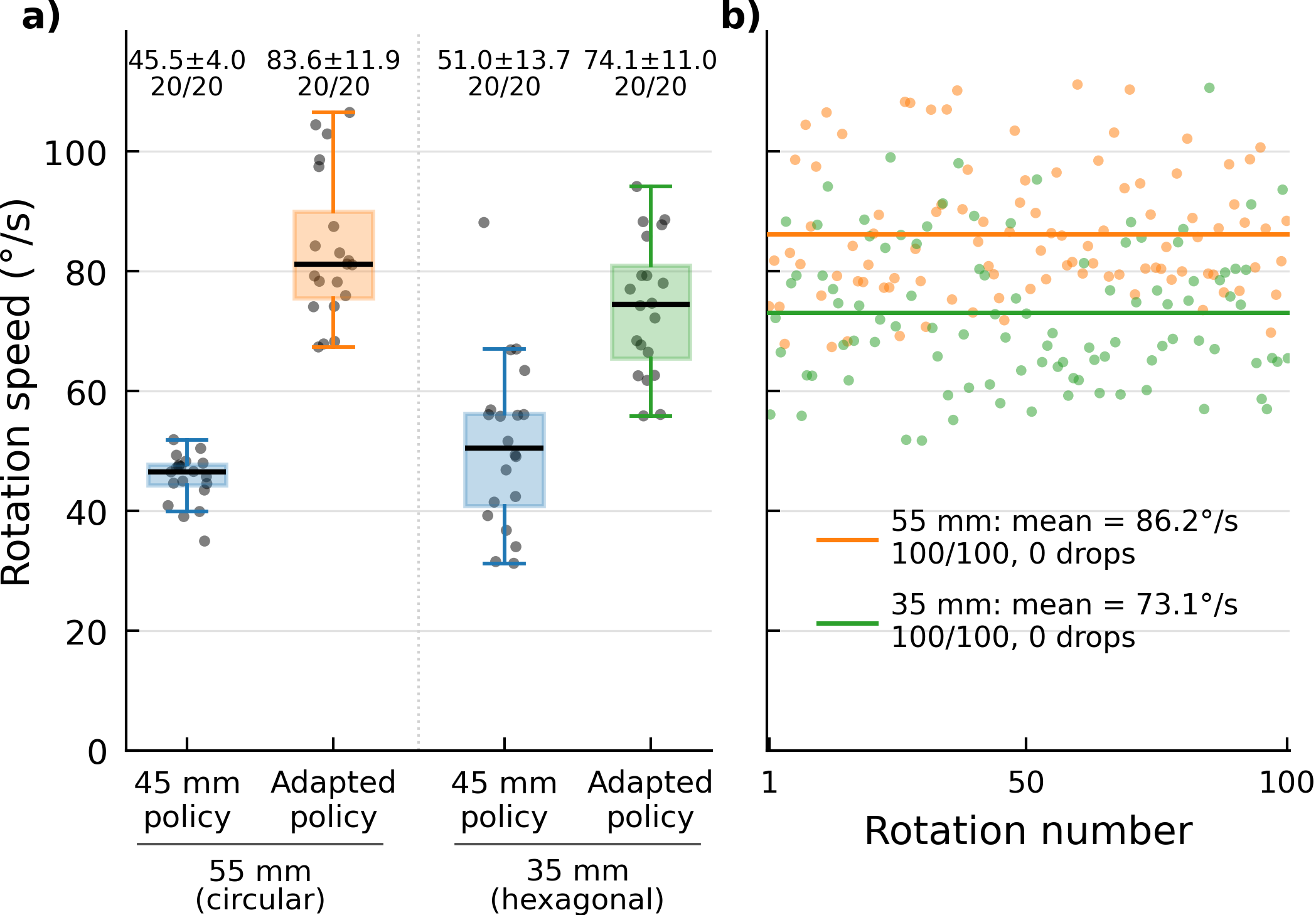}
\caption{Evaluation before and after real-world adaptation to new object geometries.}
\label{fig:adaptation_speed}
\end{figure}

% Figure~\ref{fig:adaptation_speed}(a) shows that adaptation substantially improves rotation speed on both new object geometries. On the 55~mm circular object, adaptation increases the mean rotation speed from $45.5 \pm 4.0^\circ$/s under direct transfer to $83.6 \pm 11.9^\circ$/s, corresponding to an approximately 84\% improvement. On the 35~mm hexagonal object, the mean rotation speed increases from $51.0 \pm 13.7^\circ$/s to $74.1 \pm 11.0^\circ$/s, an improvement of approximately 45\%. All four matched evaluations complete 20/20 rotations without an object drop, indicating that these substantial speed gains are achieved while maintaining reliable execution.

% To evaluate whether this improved performance is sustained over longer repeated operation, Fig.~\ref{fig:adaptation_speed}(b) shows 100 consecutive rotations for each adapted policy. Both policies complete all 100 rotations without an object drop, while maintaining mean rotation speeds of $86.2^\circ$/s for the 55~mm circular object and $73.1^\circ$/s for the 35~mm hexagonal object.

\claim{Adaptation makes the policy faster without sacrificing reliability.} Fig.~\ref{fig:adaptation_speed}(a) compares rotation speed before and after adaptation. On the 55~mm circular object, mean speed increases from $45.5 \pm 4.0^\circ$/s under direct transfer to $83.6 \pm 11.9^\circ$/s (approximately $84\%$); on the 35~mm hexagonal object, from $51.0 \pm 13.7^\circ$/s to $74.1 \pm 11.0^\circ$/s (approximately $45\%$). All four matched evaluations complete $20/20$ rotations without a drop, and over $100$ consecutive rotations (Fig.~\ref{fig:adaptation_speed}(b)) both adapted policies again complete every rotation without a drop, at mean speeds of $86.2^\circ$/s and $73.1^\circ$/s, respectively (see supplementary video).

% These experiments show that the policy learned for the nominal object already provides a useful initialization across moderate changes in object size and cross-section. Additional real-world learning then specializes the policy to the new geometry and substantially improves its execution speed.

These results show that the policy learned on the nominal object transfers effectively across the tested changes in object size and cross-section and provides a useful starting point for further specialization. Rather than recollecting MPC trajectories and repeating pretraining for each new object, continued real-world learning with online MPC guidance improves execution speed while retaining reliable rotation.

\subsection{Goal-Conditioned Reorientation}
\label{sec:reorientation}

% We evaluate the proposed learning approach on a goal-conditioned reorientation task. Unlike continuous rotation, where the object is rotated about its principal axis, this task requires reorientation about a perpendicular axis. The policy receives a desired object heading and rotates the object toward the commanded target while maintaining the grasp. During training and evaluation, the desired heading is randomly varied within a $\pm20^\circ$ range.

We further evaluate the framework on goal-conditioned in-hand reorientation about an axis perpendicular to the object's principal axis (Fig.~\ref{fig:exp_setup}(b)). Whereas the continuous rotation task requires sustained manipulation about the principal axis in a fixed rotational direction, this task requires the policy to reach commanded object headings while maintaining the grasp. Together, the two tasks evaluate the framework on both repeated cyclic manipulation and target-directed reorientation. The purpose of this experiment is therefore to test whether the same use of MPC for prior experience, pretraining, and temporary online guidance extends to a different goal-conditioned dexterous manipulation objective. During training and evaluation, the desired heading is randomly varied within a $\pm20^\circ$ range.

% The hand-control action space is unchanged from the continuous-rotation task, but the policy is additionally conditioned on the desired object heading. The observation therefore contains the same robot and object state information together with the goal heading. The reward follows the same structure as Eq.~\ref{eq:rotation_reward}, with the rotation-progress term replaced by progress toward the commanded heading, measured as the reduction in absolute heading error between consecutive steps. The reward otherwise retains the same time penalty, success bonus, and terminal drop penalty.

The hand-control action space is unchanged from the continuous-rotation task, but the policy is additionally conditioned on the desired object heading. The observation therefore contains the same robot and object state information together with the goal heading. The reward follows the same structure as Eq.~\ref{eq:rotation_reward}, with the rotation-progress term replaced by progress toward the commanded heading, measured as the reduction in absolute heading error between consecutive steps. The reward otherwise retains the same time penalty, success bonus, and terminal drop penalty. A reorientation trial is considered successful when the absolute heading error remains within $5^\circ$ of the commanded heading for at least $0.5$\,s. Before online learning, we collect approximately 7 minutes of real-world interaction using the MPC controller and use these trajectories as the offline initialization data for replay and actor-critic pretraining described in Sec.~II.

% Figure~\ref{fig:reorientation} summarizes the real-world training process and final policy evaluation. During training, the mean final reaching error decreases rapidly from approximately $17^\circ$ in the first evaluation to only a few degrees thereafter. Over the 45-minute training period, the robot accumulates 20 object drops.

\begin{figure}[t!]
\centering
\includegraphics[width = 1.0\columnwidth]
{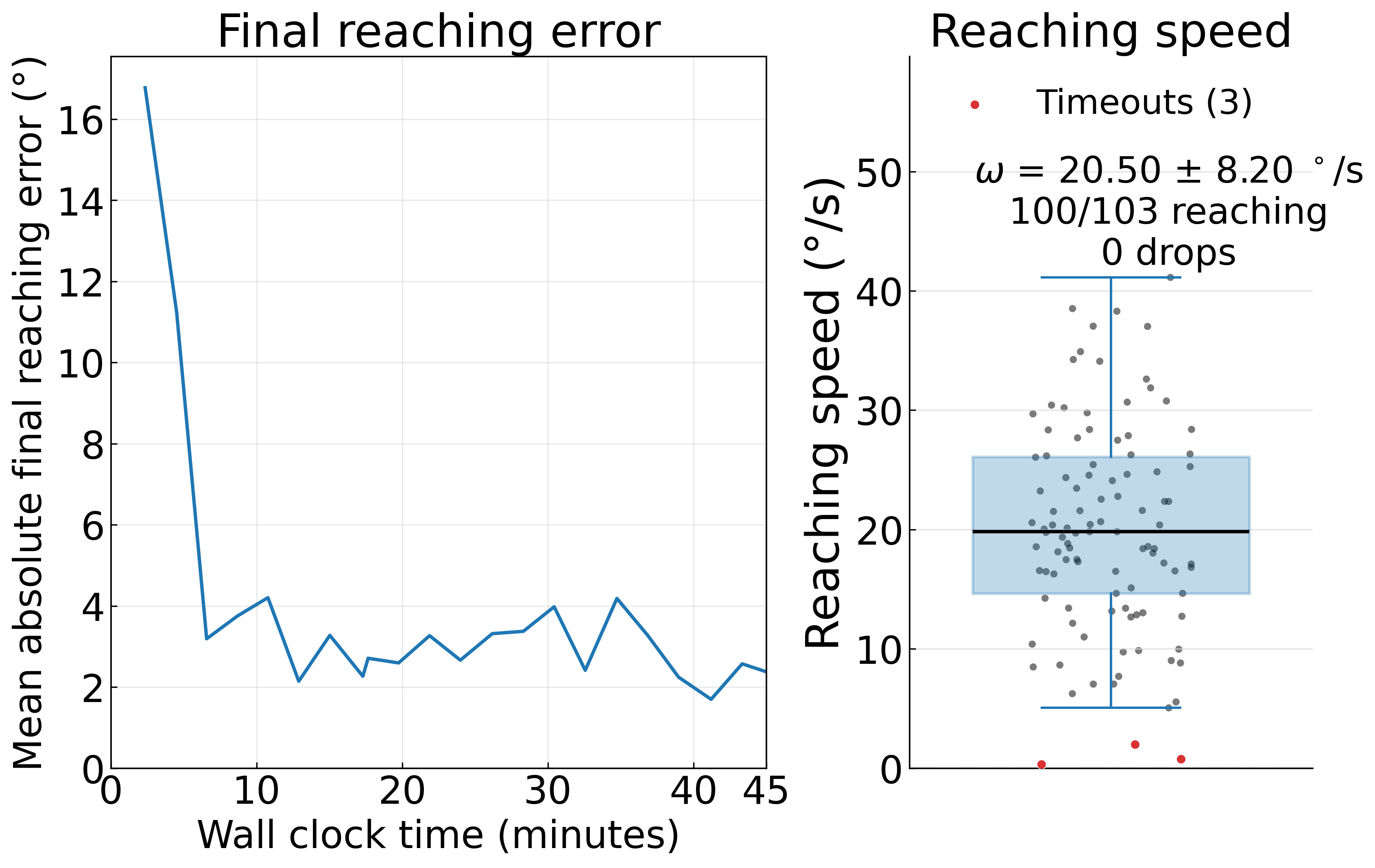}
\caption{Goal-conditioned reorientation results. (a) Mean absolute final reaching error during real-world online learning. (b) Reaching speed of the final policy over 103 evaluation trials.}
\label{fig:reorientation}
\end{figure}

Fig.~\ref{fig:reorientation} summarizes the real-world training process and final policy evaluation. During training, the mean final reaching error decreases rapidly from approximately $17^\circ$ in the first evaluation to only a few degrees thereafter. Over the 45-minute online training period, the robot accumulates 20 object drops.

% We evaluate the final policy until 100 successful goal-conditioned reorientation trials are completed, resulting in 103 total trials. The remaining three trials terminate after the 10\,s task timeout, and no object drops occur during evaluation. Across the 100 successful trials, the policy achieves a mean reaching speed of $20.50 \pm 8.20^\circ$/s.

We evaluate the final policy until 100 successful trials are completed, requiring 103 trials in total and corresponding to a $97.1\%$ success rate. The remaining three trials terminate after the 10\,s task timeout, and no object drops occur during evaluation. Across the 100 successful trials, the policy achieves a mean reaching speed of $20.50 \pm 8.20^\circ$/s.

% Unlike continuous rotation, the purpose of this experiment is not to repeat the complete baseline and ablation study. Rather, it demonstrates that the same use of MPC for initialization and early online guidance can be applied to a goal-conditioned dexterous manipulation objective rather than being specific to continuous rotation.

\section{Conclusion}
\label{sec:conclusion}

We presented a real-world reinforcement learning framework that uses sampling-based MPC as scaffolding, providing prior experience and online guidance for multi-fingered dexterous manipulation. By combining MPC replay, actor-critic pretraining, and online MPC guidance, the proposed method learns reliable continuous in-hand rotation within minutes while requiring few physical resets. The learned policy substantially outperforms both direct MPC execution and the sim-to-real PPO policy in rotation speed and completes 1000 consecutive rotations in around 110 minutes without a drop. Ablations show that replay, pretraining, and online guidance provide complementary benefits, while experiments on new object geometries and goal-conditioned reorientation demonstrate policy adaptation and applicability to different dexterous manipulation objectives. Together, these results show that MPC can serve as scaffolding for real-world dexterous RL: its experience supports replay and actor-critic initialization, while temporary online guidance structures early interaction, allowing the final policy to execute independently and outperform the planner that guided it. Future work will extend the framework from hand-only control to coordinated arm-hand manipulation and investigate longer-horizon tasks that combine arm-level object motion with dexterous in-hand control.

\addtolength{\textheight}{-0cm}   % This command serves to balance the column lengths
                                  % on the last page of the document manually. It shortens
                                  % the textheight of the last page by a suitable amount.
                                  % This command does not take effect until the next page
                                  % so it should come on the page before the last. Make
                                  % sure that you do not shorten the textheight too much.

%%%%%%%%%%%%%%%%%%%%%%%%%%%%%%%%%%%%%%%%%%%%%%%%%%%%%%%%%%%%%%%%%%%%%%%%%%%%%%%%
%\input{sections/Appendix}
% \input{sections/Acknowledgment}
%%%%%%%%%%%%%%%%%%%%%%%%%%%%%%%%%%%%%%%%%%%%%%%%%%%%%%%%%%%%%%%%%%%%%%%%%%%%%%%%
\bibliographystyle{bibliography/myIEEEtran} 
\bibliography{bibliography/references}
%%%%%%%%%%%%%%%%%%%%%%%%%%%%%%%%%%%%%%%%%%%%%%%%%%%%%%%%%%%%%%%%%%%%%%%%%%%%%%%%

\end{document}